\documentclass[11pt,a4paper]{article}
\usepackage[hyperref]{acl2018}
\usepackage{times}
\usepackage{latexsym}

\usepackage{url}

\usepackage{xcolor}

\usepackage{xspace}
\newcommand{\RE}{\texttt{RE}\xspace}
\newcommand{\NN}{\texttt{NN}\xspace}
\newcommand{\NNs}{\texttt{NNs}\xspace}

\newcommand{\SLU}{\texttt{SLU}\xspace}
\newcommand{\NLP}{\texttt{NLP}\xspace}

\newcommand{\FOL}{\texttt{FOL}\xspace}
\newcommand{\BLSTM}{\texttt{BLSTM}\xspace}
\newcommand{\LL}{\texttt{L\&L16}\xspace}
\newcommand{\tatt}{\texttt{two}\xspace}
\newcommand{\ptatt}{\texttt{+two}\xspace}

\usepackage{mathrsfs}
\usepackage{amsmath}
\usepackage{graphicx}
\usepackage{multirow}
\usepackage{makecell}
\usepackage[english, american]{babel}
\usepackage{xspace}
\usepackage{color, colortbl}

\usepackage{balance}
\usepackage{blindtext}
\usepackage{amsmath,amssymb,amsfonts}
\usepackage{algorithmic}
\usepackage{booktabs}
\usepackage{subfigure}

\usepackage{tikz}
\newcommand*\circled[1]{\tikz[baseline=(char.base)]{
            \node[shape=circle,draw,inner sep=0.1pt] (char) {#1};}}

\newcommand\REs{\texttt{REs}\xspace}
\newcommand\REO{\texttt{REO}\xspace}

\newcommand\REtag{\texttt{REtag}\xspace}
\newcommand\REtags{\texttt{REtags}\xspace}
\newcommand\BIO{\texttt{BIO}\xspace}

\definecolor{Gray}{gray}{0.9}

\newboolean{showcomments}
\setboolean{showcomments}{true} 
\ifthenelse{\boolean{showcomments}} { \newcommand{\mynote}[3]{
    \fbox{\bfseries\sffamily\scriptsize#1}
    {\small$\blacktriangleright$\textsf{\emph{\color{#3}{#2}}}$\blacktriangleleft$}}}
{ \newcommand{\mynote}[3]{}}
\newcommand{\shrink}[1]{}

\newcommand{\tabincell}[2]{\begin{tabular}{@{}#1@{}}#2\end{tabular}}

\newcommand{\cparagraph}[1]{\vspace{1.5mm}\noindent\textbf{#1.}}

\DeclareMathOperator{\softmax}{softmax}

\aclfinalcopy 
\def\aclpaperid{1500} 

\title{Marrying Up Regular Expressions with Neural Networks: \\A Case Study for Spoken Language Understanding}

\author{Bingfeng Luo$^1$, Yansong Feng$^{*1}$, Zheng Wang$^2$,
	\\\textbf{Songfang Huang$^3$, Rui Yan$^1$ \and Dongyan Zhao$^1$}\\
	$^1$ICST, Peking University, China\\
	$^2$MetaLab, Lancaster University, UK \\
	$^3$IBM China Research Lab, China \\
	{\tt \{bf\_luo,fengyansong,ruiyan,zhaody\}@pku.edu.cn} \\
	{\tt z.wang@lancaster.ac.uk, huangsf@cn.ibm.com} \\}

\date{}

\begin{document}
\maketitle

\begin{abstract}
The success of many natural language processing (\NLP) tasks is bound by the number and quality of annotated data, but there is often a
shortage of such training data. In this paper, we ask the question: ``Can we combine a neural network (\NN) with regular expressions
(\RE) to improve supervised learning for \NLP?". In answer, we develop novel methods to exploit the rich expressiveness of \REs at
different levels within a \NN, showing that the combination significantly enhances the learning effectiveness when a small number of
training examples are available. We evaluate our approach by applying it to spoken language understanding for intent detection and slot
filling. Experimental results show that our approach is highly effective in exploiting the available training data, giving a clear boost
to the \RE-unaware \NN.



\end{abstract}

\section{Introduction}

Regular expressions (\REs) are widely used in various natural language processing (\NLP) tasks like pattern matching, sentence
classification, sequence labeling, etc.~\cite{chang2014tokensregex}.
As a technique based on human-crafted rules, it is concise, interpretable, tunable, and does not rely on much training data to generate. As
such, it is commonly used in industry, especially when the available training examples are limited -- a problem known as few-shot
learning~\cite{gc2015big}.

While powerful, \REs have a poor generalization ability because all synonyms and variations in a \RE must be explicitly specified. As a
result, 
\REs are often ensembled  with data-driven methods, such as neural network (\NN) based techniques, where a set of carefully-written \REs
are used to handle certain cases with high precision, leaving the rest for data-driven methods.

We believe the use of \REs can go beyond simple pattern matching. In addition to being a separate classifier to be ensembled, a \RE also
encodes a developer's knowledge for the problem domain. The knowledge could be, for example,  the informative words (\textbf{\textit{clue
words}}) within a \RE's surface form. We argue that such information can be utilized by data-driven methods to achieve better prediction
results, especially in few-shot learning.

This work investigates the use of \REs to improve \texttt{NNs} -- a learning framework that is widely used in many \NLP
tasks~\cite{goldberg2017neural}. The combination of \REs and a \NN  allows us to exploit the conciseness and effectiveness of \REs and the
strong generalization ability of \NNs. This also provides us  an opportunity to learn from various kinds of  \REs, since \NNs are known to
be good at tolerating noises~\cite{xie2016disturblabel}.

This paper presents novel approaches to combine \REs with a \NN at different levels.  At the input layer, we propose to use the evaluation
outcome of \REs as the input features of a \NN (Sec.\ref{fusion_with_input}). At the network module level, we show how to exploit the
knowledge encoded in \REs to guide the attention mechanism of a \NN (Sec.~\ref{interact_with_module}). At the output layer, we combine the
evaluation outcome of a \RE with the \NN output in a learnable manner (Sec.~\ref{fusion_with_output}).

%
We evaluate our approach by applying it to two spoken language understanding (\SLU) tasks, namely \emph{intent detection} and \emph{slot
filling}, which respectively correspond to two fundamental \NLP tasks: sentence classification and sequence labeling. To demonstrate the
usefulness of \REs in real-world scenarios where the available number of annotated data can vary, we explore both the few-shot learning
setting and the one with full training data. Experimental results show that our approach is highly effective in utilizing the available
annotated data, yielding significantly better learning performance over the \RE-unaware method.



Our contributions are as follows. (1) We present the first work to systematically investigate methods for combining \REs with \NNs. (2) The
proposed methods are shown to clearly improve the \NN performance in both the few-shot learning and the full annotation settings. (3) We
provide a set of guidance on how to combine \REs with \NNs and \RE annotation.

\section{Background}
\begin{figure}[t!]
  \centering
  \includegraphics[width=0.49\textwidth]{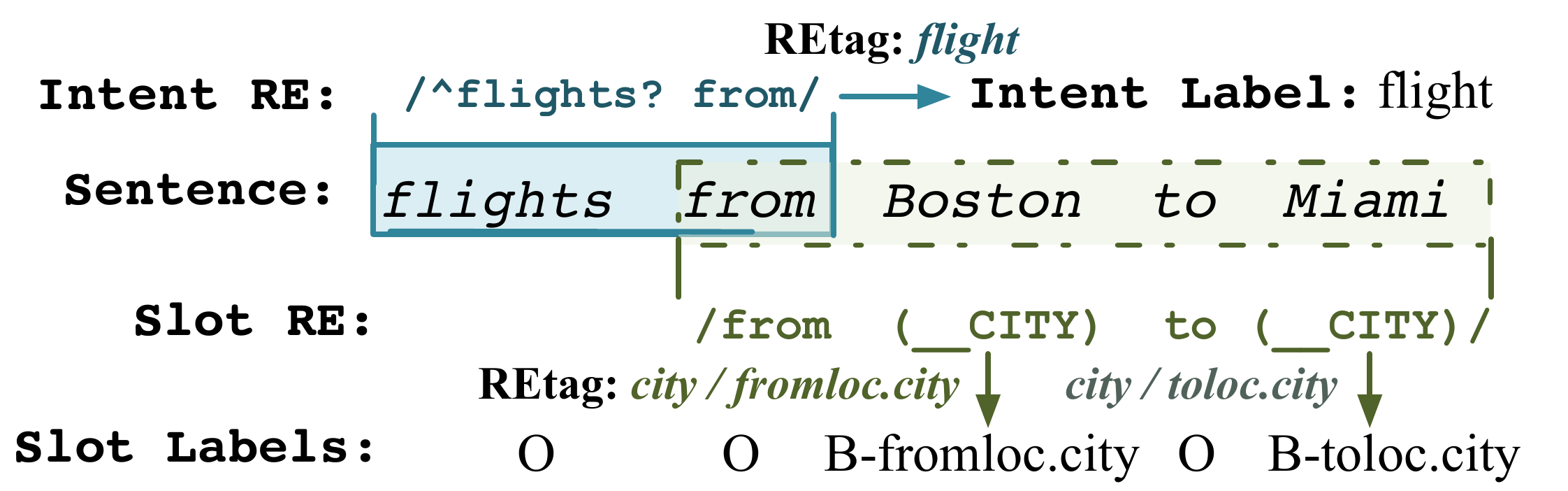}\\
  \vspace{-3mm}
  \caption{A sentence from the ATIS dataset. \REs can be used to detect the intent and label slots.}
  \label{atis_sample}
  \vspace{-3mm}
\end{figure}

\vspace{-1mm}
\subsection{Typesetting}
\vspace{-1mm} In this paper, we use italic for emphasis like \emph{intent detection}, the Courier typeface for abbreviations like
\texttt{RE}, bold italic for the first appearance of a concept like \emph{\textbf{clue words}}, Courier surrounded by / for regular
expressions like {\small \texttt{/list(\;the)?\;\_\_AIRLINE/}}, and underlined italic for words of sentences in our dataset like
\underline{\textit{Boston}}.

\vspace{-1mm}
\subsection{Problem Definition}
\vspace{-1mm}
Our work targets two \SLU tasks: \emph{intent detection} and \emph{slot filling}. The former is a sentence classification task where we
learn a function to map an input sentence of $n$ words, $\textbf{x}=[x_{1}, ..., x_{n}]$, to a corresponding \textbf{\emph{intent label}},
$c$. The latter is a sequence labeling task for which we learn a function to take in an input query sentence of $n$ words,
$\textbf{x}=[x_{1}, ..., x_{n}]$, to produce a corresponding labeling sequence, $\textbf{y}=[y_{1}, ..., y_{n}]$, where  $y_i$ is the
\textbf{\emph{slot label}} of the
corresponding word, $x_i$. 

%


Take the sentence in Fig.~\ref{atis_sample} as an example.
A successful intent detector would suggest the intent of the sentence as \emph{flight}, i.e., querying about flight-related information. A
slot filler, on the other hand, should identify the slots \emph{fromloc.city} and \emph{toloc.city} by labeling \underline{\textit{Boston}}
and \underline{\textit{Miami}}, respectively,
using the begin-inside-outside (\texttt{BIO}) scheme.

\subsection{The Use of Regular Expressions}
\label{re_desc} \vspace{-1mm}

In this work, a \RE defines a mapping from a text \emph{pattern} to several \textbf{\emph{\REtags}} which are  the same as or related to
the \textbf{\emph{target labels}} (i.e., intent and slot labels). A search function takes in a \RE, applies it to all sentences, and
returns any texts that match the pattern. We then assign the \REtag(\texttt{s}) (that are associated with the matching \RE) to either the
matched sentence (for intent detection) or some matched phrases (for slot filling).

Specifically, our \REtags for intent detection are the same as the intent labels.
For example,  
in Fig.~\ref{atis_sample},
we get a \REtag of \emph{flight} that is the same as 
 the intent label \emph{flight}.


For slot filling, we use two different sets of \REs. Given the group functionality of \RE, we can assign \REtags to our interested
\textbf{\emph{\RE groups}} (i.e., the expressions defined inside parentheses). The translation from \REtags to slot labels depends on how
the corresponding \REs are used. (1) When \REs are used at the network module level (Sec.~\ref{interact_with_module}), the corresponding
\REtags are the same as the target slot labels. For instance, the slot \RE in Fig.~\ref{atis_sample} will assign \emph{fromloc.city} to the
first \RE group and \emph{toloc.city} to the second one. Here,  {\small \texttt{\_\_CITY}} is a list of city names, which can be replaced
with a \RE string like \texttt{\small/Boston|Miami|LA|.../}. (2) If \REs are used in the input (Sec.~\ref{fusion_with_input}) and the
output layers (Sec.~\ref{fusion_with_output}) of a \NN, the corresponding \REtag would be different from the target slot labels. In this
context, the two \RE groups in Fig.~\ref{atis_sample} would be simply tagged as \emph{city} to capture the commonality of three related
target slot labels: \emph{fromloc.city}, \emph{toloc.city}, \emph{stoploc.city}. Note that we could use the target slot labels as \REtags
for all the settings. The purpose of abstracting \REtags to a simplified version of the target slot labels here is to show that \REs can
still be useful when their evaluation outcome does not exactly match our learning objective. Further, as shown in Sec.~\ref{re_in_exp},
using simplified \REtags can also make the development of \REs easier in our tasks.
Intuitively, complicated \REs can lead to better performance but require more efforts to generate. 
 Generally, there are two aspects affecting \RE complexity most: the number of \RE groups\footnote{
	When discussing complexity, we consider each semantically independent consecutive word sequence as a \RE group (excluding clauses, such as \texttt{\textbackslash w+}, that can match any word).
	For instance, the \texttt{RE}: {\small \texttt{/how\:long(\:\textbackslash w+)\{1,2\}?\:(it\:take|flight)/}} has two RE groups: {\small \texttt{(how\;long)}} and {\small \texttt{(it\;take|flight)}}.
}
and \emph{or} clauses (i.e., expressions separated by the disjunction operator $|$) in a \RE group. Having a larger number of \RE groups
often leads to better precision but lower coverage on pattern matching, while a larger number of \emph{or} clauses usually gives a higher
coverage but slightly lower precision.



\begin{figure*}[!t]
\centering
\subfigure[Intent Detection] {
    \label{fig_overview_intent}
    \includegraphics[width=0.8\columnwidth]{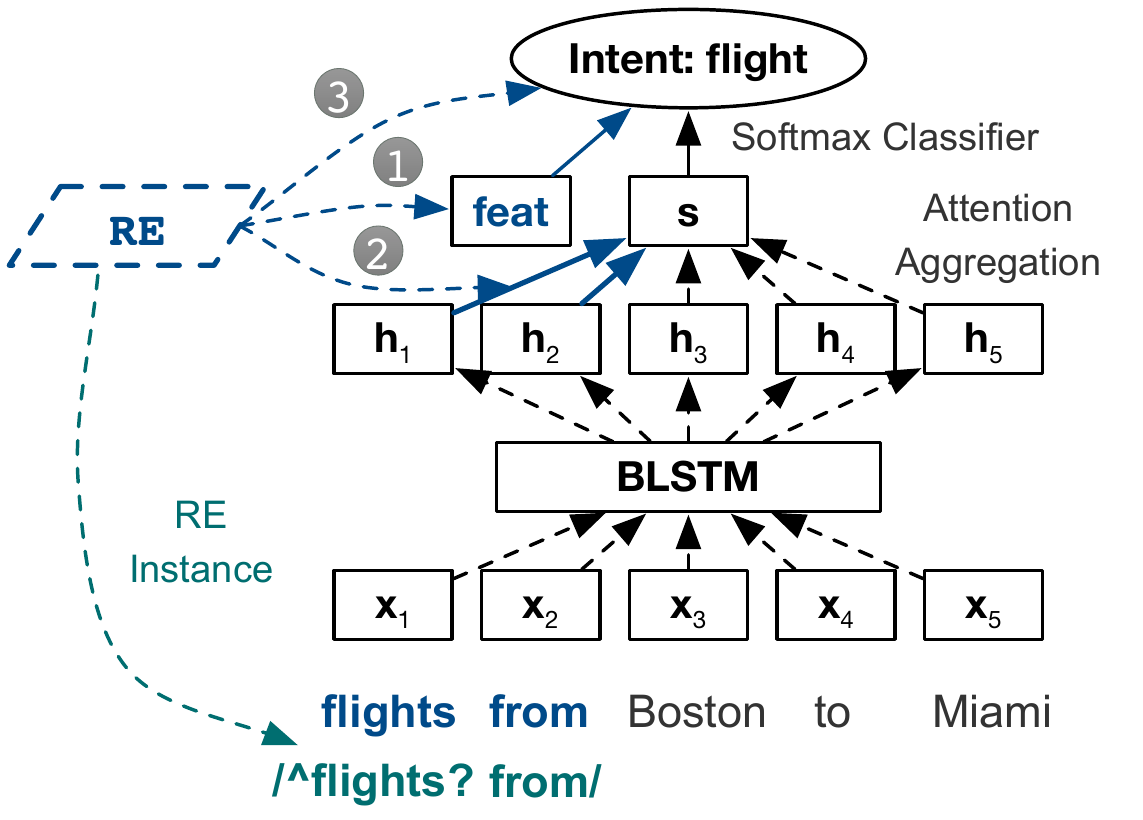}
}
\hspace{.5in}
\subfigure[Slot Filling (predicting slot label for \textsl{\underline{Boston}})] {
    \label{fig_overview_slot}
    \includegraphics[width=0.8\columnwidth]{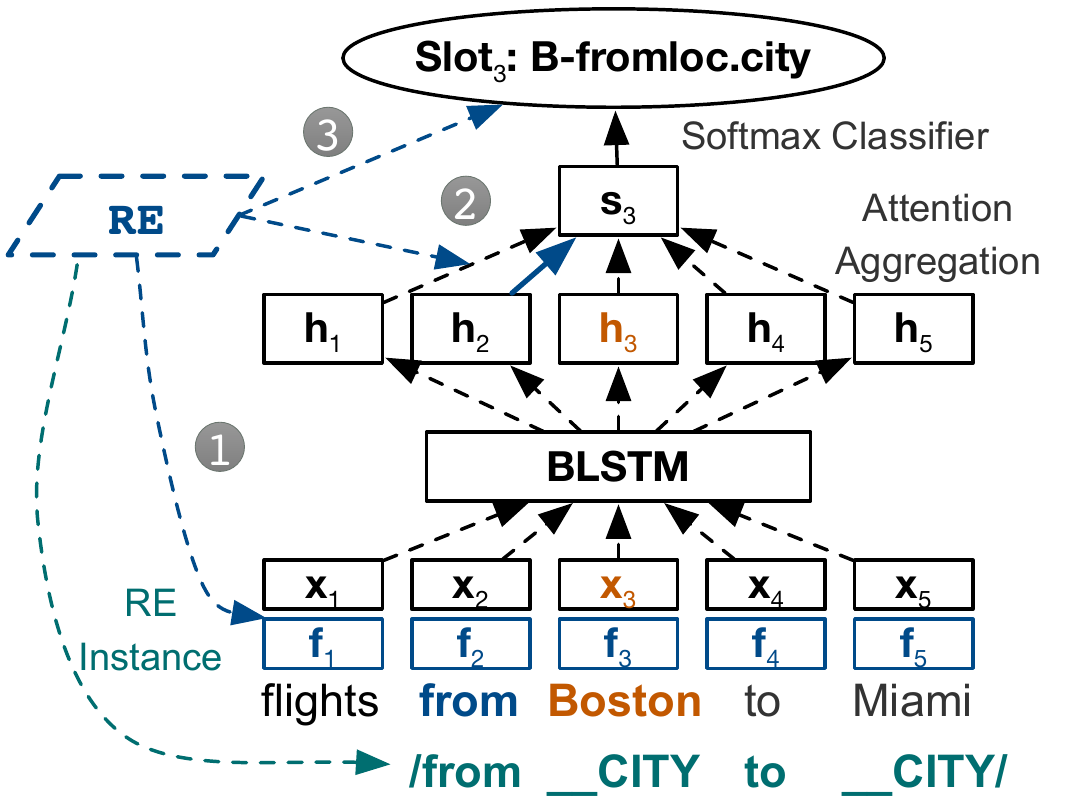}
}
\vspace{-3mm}
\caption{Overview of our methods. \circled{1}, \circled{2}, \circled{3} refers to the methods in
Sec.~\ref{fusion_with_input}, \ref{interact_with_module}, \ref{fusion_with_output} respectively.}
\label{fig_overview}
\vspace{-5mm}
\end{figure*}

\section{Our Approach}
As depicted in Fig.~\ref{fig_overview}, we propose  to combine \NNs and \REs from three different angles.

\subsection{Base Models}
\label{sec:baseline} We use the Bi-directional LSTM (\BLSTM) as our base \NN model because it is effective in both intent detection and
slot filling~\cite{liu2016attention}.

\cparagraph{Intent Detection} As shown in Fig.~\ref{fig_overview}, the \BLSTM takes as input the word embeddings $[\textbf{x}_1, ...,
\textbf{x}_n]$ of a n-word sentence, and produces a vector $\textbf{h}_i$ for each word $i$. A self-attention layer then takes in the
vectors produced by the \BLSTM to compute the sentence embedding $\textbf{s}$:
\begin{equation}
\textbf{s} = \sum_{i}{\alpha_i\textbf{h}_i}, \quad \alpha_i=\frac{\exp(\textbf{h}_i^\intercal \textbf{Wc})}{\sum_{i}{\exp(\textbf{h}_i^\intercal \textbf{Wc})}}
\label{eq:simple_att}
\end{equation}
where  $\alpha_i$ is the attention for word $i$, $\textbf{c}$ is a randomly initialized trainable vector used to select informative words for classification, and $\textbf{W}$ is a weight matrix.
Finally, $\textbf{s}$ is fed to a softmax classifier for intent classification.

\cparagraph{Slot Filling} The model for slot filling is  straightforward -- the slot label prediction is generated by a softmax classier
which takes in the \BLSTM's output $\textbf{h}_i$ and produces the slot label of word $i$. Note that attention aggregation in
Fig.~\ref{fig_overview} is only employed by the network module level method presented in Sec.~\ref{interact_with_module}.

\subsection{Using REs at the Input Level}
\label{fusion_with_input}
At the input level, we use the evaluation outcomes of \REs as features which are fed to \NN models.

\cparagraph{Intent Detection}
Our \REtag for intent detection is the same as our target intent label.
Because real-world \REs are unlikely to be perfect, one sentence may be matched by more than one \RE. This may result in several \REtags
that are conflict with each other. For instance, the sentence \textsl{\underline{list the Delta airlines flights to Miami}} can match a
\RE: {\small \texttt{/list(\;the)?\;\_\_AIRLINE/}} that outputs tag \emph{airline}, and another \RE: {\small \texttt{/list(\,\textbackslash
w+)\{0,3\} flights?/}} that outputs tag \emph{flight}.

To resolve the conflicting situations illustrated above, we average the randomly initialized trainable tag embeddings to form an aggregated
embedding as the \NN input. There are two ways to use the aggregated embedding. We can  append the aggregated embedding to either the
embedding of every input word, or the input of the softmax classifier (see \circled{1} in Fig.~\ref{fig_overview_intent}). To determine
which strategy works best, we perform a pilot study. We found that the first method causes the tag embedding to be copied many times;
consequently, the \NN tends to heavily rely on the \REtags, and the resulting performance is similar to the one given by using \REs alone
in few-shot settings. Thus, we adopt the second approach.

\cparagraph{Slot Filling} Since the evaluation outcomes of slot \REs are word-level tags,
we can simply embed and average the \REtags into a vector $\textbf{f}_i$ for each word, and append it
to the corresponding word embedding $\textbf{w}_i$ (as shown in \circled{1} in Fig.~\ref{fig_overview_slot}).
Note that we also extend the slot \REtags into the \BIO format, e.g., the \REtags of phrase \textsl{\underline{New York}} are \emph{B-city} and \emph{I-city} if its original tag is \emph{city}.

\subsection{Using REs at the Network Module Level}
\label{interact_with_module} At the network module level, we explore ways to utilize the clue words in the surface form of a \RE (bold blue arrows and
words in \circled{2} of Fig.~\ref{fig_overview}) to guide the attention module in \NNs.


\cparagraph{Intent Detection} Taking the sentence in Fig.~\ref{atis_sample} for example, the \RE: {\small\texttt{/\textasciicircum
flights?\:from/} } that leads to intent \emph{flight} means that \textsl{\underline{flights from}} are the key words to decide the intent
\emph{flight}. Therefore, the attention module in \NNs should leverage these two words to get the correct prediction. To this end, we
extend the base intent model by making two changes to incorporate the guidance from \REs.

First, since each intent has its own clue words, using a single sentence embedding for all intent labels 
would make the attention less focused.
Therefore, we let each intent label $k$ use different attention $\textbf{a}_k$, which is then used to generate the sentence embedding
$\textbf{s}_k$ for that intent:
\begin{equation}
\textbf{s}_k = \sum_{i}{\alpha_{ki}\textbf{h}_i}, \quad
\alpha_{ki}=\frac{\exp(\textbf{h}_i^\intercal \textbf{W}_a\textbf{c}_k)}{\sum_{i}{\exp(\textbf{h}_i^\intercal \textbf{W}_a\textbf{c}_k)}}
\label{label_att_eq}
\end{equation}
where $\textbf{c}_k$ is a trainable vector for intent $k$ which is used to compute attention $\textbf{a}_k$, $\textbf{h}_i$ is the \BLSTM output for word $i$, and $\textbf{W}_a$ is a weight matrix.

The probability $p_k$ that the input sentence expresses intent $k$ is computed by:
\begin{equation}
p_k = \frac{\exp(logit_k)}{\sum_{k}{\exp(logit_k)}}, \quad\quad logit_k=\textbf{w}_k\textbf{s}_k + b_k
\label{label_prob_eq}
\end{equation}
where $\textbf{w}_k$, $logit_k$, $b_k$ are weight vector, logit, and bias for intent $k$, respectively.

Second, apart from indicating a sentence for intent $k$ (\textbf{\emph{positive \REs}}),
a \RE can also indicate that a sentence does not express intent $k$ (\textbf{\emph{negative \REs}}).
We thus use a new set of attention (\textbf{\emph{negative attentions}}, in contrast to \textbf{\emph{positive attentions}}), to compute
another set of logits for each intent with Eqs.~\ref{label_att_eq} and \ref{label_prob_eq}. We denote the logits computed by positive
attentions as $logit_{pk}$, and those by negative attentions as $logit_{nk}$, the final logit for intent $k$ can then be calculated as:
\begin{equation}
logit_k = logit_{pk} - logit_{nk}
\end{equation}

To use \REs to guide attention, we add an attention loss to the final loss:
\begin{equation}
loss_{att} = \sum_{k}{\sum_{i}{t_{ki}\log(\alpha_{ki})}}
\label{att_loss}
\end{equation}
where $t_{ki}$ is set to $0$ when none of the matched \REs (that leads to intent $k$) marks word $i$ as a clue word -- otherwise $t_{ki}$
is set to $1/l_{k}$, where $l_k$ is the number of clue words
for intent $k$ (if no matched \RE leads to intent $k$, then $t_{k*}=0$).
We use Eq.~\ref{att_loss}
to compute the positive attention loss, $loss_{att\_p}$, for positive \REs and negative attention loss, $loss_{att\_n}$, for negative ones.
The final loss is computed as:
\begin{equation}
loss = loss_{c} + \beta_p loss_{att\_p} + \beta_n loss_{att\_n}
\end{equation}
where $loss_{c}$ is the original classification loss, $\beta_p$ and $\beta_n$ are weights for the two attention losses.

\cparagraph{Slot Filling}
The \textbf{\emph{two-side attention}} (positive and negative attention) mechanism introduced for intent prediction is unsuitable for slot
filling. Because for slot filling, we need to compute attention for each word, which demands more computational and memory resources than
doing that for intent detection\footnote{Since we need to assign a label to each word, if we still compute attention for each slot label,
we will have to compute $2\times L \times n^2$ attention values for one sentence. Here, $L$ is the number of tags and $n$ is the sentence
length. The \BIO tagging format will further double the number of tags.}.

Because of the aforementioned reason, we use a simplified version of the two-side attention, where all the slot labels share the same set
of positive and negative attention. Specifically, to predict the slot label of word $i$, we use the following equations, which are similar
to Eq.~\ref{eq:simple_att}, to generate a sentence embedding $\textbf{s}_{pi}$ with regard to word $i$ from positive attention:
\begin{equation}
\textbf{s}_{pi} = \sum_{j}{\alpha_{pij}\textbf{h}_j}, \quad \alpha_{pij}=\frac{\exp(\textbf{h}_j^\intercal\textbf{W}_{sp}\textbf{h}_i)}{\sum_{j}{\exp(\textbf{h}_j^\intercal \textbf{W}_{sp}\textbf{h}_i)}}
\label{eq:slu_simple_att}
\end{equation}
where $\textbf{h}_i$ and $\textbf{h}_j$ are the \BLSTM outputs for word $i$ and $j$ respectively, $\textbf{W}_{sp}$ is a weight matrix, and
$\alpha_{pij}$ is the positive attention value for word $j$ with respect to word $i$. Further, by replacing $\textbf{W}_{sp}$ with
$\textbf{W}_{sn}$, we use  Eq.~\ref{eq:slu_simple_att} again to compute negative attention and generate the corresponding sentence
embedding $\textbf{s}_{ni}$.

Finally, the prediction $\textbf{p}_i$ for word $i$ can be calculated as:
\begin{equation}
\begin{split}
\textbf{p}_i = \softmax((\textbf{W}_p [\textbf{s}_{pi}; \textbf{h}_i] + \textbf{b}_p) \\- (\textbf{W}_n [\textbf{s}_{ni}; \textbf{h}_i] + \textbf{b}_n))
\end{split}
\end{equation}
where $\textbf{W}_{p}$, $\textbf{W}_{n}$, $\textbf{b}_{p}$, $\textbf{b}_{n}$ are weight matrices and bias vectors for positive and negative attention, respectively. Here we append the \BLSTM output $\textbf{h}_i$ to $\textbf{s}_{pi}$ and $\textbf{s}_{ni}$ because the word $i$ itself also plays a crucial part in identifying its slot label.

\subsection{Using REs at the Output Level}
\label{fusion_with_output} At the output level, \REs are used to amend the output of \NNs. At this level, we take the same approach used
for intent detection and slot filling (see \circled{3} in Fig.~\ref{fig_overview}).


As mentioned in Sec.~\ref{re_desc}, the slot \REs used in the output level only produce a simplified version of target slot labels, for which
we can further
annotate their corresponding target slot labels.
For instance, a \RE that outputs \emph{city} can lead to three slot labels: \emph{fromloc.city}, \emph{toloc.city},
\emph{stoploc.city}.

Let $z_k$ be a 0-1 indicator of whether there is at least one matched \RE that leads to target label $k$ (intent or slot label), the final
logits of label $k$ for a sentence (or a specific word for slot filling) is:
\begin{equation}
logit_k = logit'_k + w_k z_k
\end{equation}
where $logit'_k$ is the logit produced by the original \NN, and $w_k$ is a trainable weight indicating the overall confidence for \REs that
lead to target label $k$. Here we do not assign a trainable weight for each \RE because it is often that only a few sentences match a \RE.

We modify the logit instead of the final probability because a logit is an unconstrained real value, which matches the property of $w_k
z_k$ better than probability. Actually, when performing model ensemble,
ensembling with logits is often empirically better than with the final probability\footnote{ An example can be found in the ensemble
version that Juan et al.~\shortcite{juan2016field} used in the Avazu Kaggle competition. }. This is also the reason why we choose to
operate on logits in Sec.~\ref{interact_with_module}.

\section{Evaluation Methodology}
Our experiments
aim to 
answer three questions: \textbf{Q1:} Does the use of \REs enhance the learning quality when the number of annotated
instances is small?  \textbf{Q2:}  Does the use of \REs still help when using the full training data?
 \textbf{Q3:}  How can we choose from different combination methods?

\subsection{Datasets}
\label{sec_datasest}

We use the ATIS dataset~\cite{hemphill1990atis} to evaluate our approach. This dataset is widely used in \SLU research. It includes queries
of flights, meal, etc. We follow the setup of Liu and Lane~\shortcite{liu2016attention} by using 4,978 queries for training and 893 for
testing, with 18 intent labels and 127 slot labels.
We also split words like \textsl{\underline{Miami's}} into \textsl{\underline{Miami 's}} during the tokenization phase to reduce the number
of words that do not have a pre-trained word embedding. This strategy is useful for few-shot learning.

To answer  \textbf{Q1} , we also exploit the \textbf{\emph{full few-shot learning setting}}. Specifically, for intent detection, we
randomly select 5, 10, 20 training instances for each intent to form the few-shot training set; and for slot filling, we also explore 5,
10, 20 shots settings. However, since a sentence typically contains multiple slots, the number of mentions of frequent slot labels may
inevitably exceeds the target shot count. To better approximate the target shot count, we select sentences for each slot label  in
ascending order of label frequencies. That is $k_1$-shot dataset will contain $k_2$-shot dataset if $k_1>k_2$. All settings use the
original test set.

Since most existing few-shot learning methods require either many few-shot classes or some classes with enough data for training, we also
explore the \textbf{\emph{partial few-shot learning setting}} for intent detection to provide a fair comparison for existing few-shot
learning methods. Specifically, we let the 3 most frequent intents have 300 training instances, and the rest remains untouched. This is
also a common scenario in real world, where we often have several frequent classes and many classes with limited data. As for slot filling,
however, since the number of mentions of frequent slot labels already exceeds the target shot count, the original slot filling few-shot
dataset can be directly used to train existing few-shot learning methods. Therefore, we do not distinguish full and partial few-shot
learning for slot filling.

\subsection{Preparing REs}
\label{re_in_exp} We use the syntax of \REs in Perl in this work. Our \REs are written by a paid annotator who is familiar with the domain.
It took the annotator in total less than 10 hours
to develop all the \REs, while a domain expert can accomplish the task faster.
We use the 20-shot training data to develop the \REs, but word lists
like cities are obtained from the full training set.
The development of \REs is considered completed when the \REs can cover most of the cases in the 20-shot training data with resonable precision.
After that, the \REs are fixed throughout the experiments.

The majority of the time
for 
writing the \REs is proportional to the number of \RE groups.
It took about 1.5 hours to write the 54 intent \REs with on average 2.2 groups per \RE. It is straightforward
to write the slot \REs for the input and output level methods, for which it took around
 1 hour to write the 60 \REs with 1.7 groups on average. By contrast, writing slot \REs to guide attention requires more
efforts as the annotator needs to carefully select clue words and annotate the full slot label. As a result, it took about
5.5 hours to generate 115 \REs with on average 3.3 groups.
The performance of the \REs can be found in the last line of Table~\ref{tab_full_few}.

In practice, a positive \RE for intent (or slot) $k$ can often be treated as negative \REs for other intents (or slots). As such, we use the positive \REs for intent (or slot) $k$ as the negative \REs for other intents (or slots) in our experiments.

\subsection{Experimental Setup}
\cparagraph{Hyper-parameters} Our hyper-parameters for the \BLSTM are similar to the ones used by Liu and
Lane~\shortcite{liu2016attention}. Specifically, we use batch size 16, dropout probability 0.5, and \BLSTM cell size 100. The attention
loss weight is 16 (both positive and negative) for full few-shot learning settings and 1 for other settings. We use the 100d GloVe word
vectors ~\cite{pennington2014glove} pre-trained on Wikipedia and Gigaword~\cite{parker2011english}, and the Adam
optimizer~\cite{kingma2014adam} with learning rate 0.001.

\cparagraph{Evaluation Metrics}
We report accuracy and macro-F1 for intent detection, and micro/macro-F1 for slot filling.
Micro/macro-F1 are the harmonic mean of micro/macro precision and recall.
Macro-precision/recall are calculated by averaging precision/recall of each label, and micro-precision/recall are averaged over each prediction.

\cparagraph{Competitors and Naming Conventions} Here, a bold Courier typeface like \textbf{\BLSTM} denotes the notations of the models that
we will compare in Sec.~\ref{sec:experiments}.

Specifically, we compare our methods with the baseline \textbf{\texttt{BLSTM}} model (Sec.~\ref{sec:baseline}). Since our attention loss
method (Sec.~\ref{interact_with_module}) uses two-side attention, we include the raw two-side attention model without attention loss
(\textbf{\ptatt}) for comparison as well. Besides, we also evaluate the \RE output (\textbf{\texttt{REO}}), which uses the \REtags as
prediction directly, to show the quality of the \REs that we will use in the experiments.\footnote{
	For slot filling, we evaluate the \REs that use the target slot labels as \REtags.}

As for our methods for combinging \REs with \NN,
\textbf{\texttt{+feat}}  refers to using \REtag as input features (Sec.~\ref{fusion_with_input}),
\textbf{\texttt{+posi}} and \textbf{\texttt{+neg}} refer to using positive and negative attention loss respectively,
\textbf{\texttt{+both}} refers to using both postive and negative attention losses (Sec.~\ref{interact_with_module}),
and \textbf{\texttt{+logit}} means using \REtag to modify \NN output (Sec.~\ref{fusion_with_output}).

Moverover, since the \REs can also be formatted as first-order-logic (\texttt{FOL}) rules, we also compare our methods with the
teacher-student framework proposed by Hu et al.~\shortcite{hu2016harnessing}, which is a general framework for distilling knowledge from
\texttt{FOL} rules into \NN (\textbf{\texttt{+hu16}}). Besides, since we consider few-short learning,  we also include the memory module
proposed by Kaiser et al.~\shortcite{kaiser2017learning}, which performs well in various few-shot datasets
(\textbf{\texttt{+mem}})\footnote{
	We tune $C$ and $\pi_0$ of \texttt{hu16}, and choose (0.1, 0.3) for intent, and (1, 0.3) for slot. We tune memory-size and $k$ of
\texttt{mem}, and choose (1024, 64) for intent, and (2048, 64) for slot. }. Finally, the state-of-art model on the ATIS dataset is also
included (\textbf{\LL}), which jointly models the intent detection and slot filling in a single network~\cite{liu2016attention}.


\section{Experimental Results}
\label{sec:experiments}

\subsection{Full Few-Shot Learning}
To answer \textbf{Q1} , we first explore the full few-shot learning scenario.

\begin{table*}
\setlength{\tabcolsep}{0.23em}
\centering
\small{
\begin{tabular}{|l|l|c|c|c|c|c|c|}

\hline
\multirow{3}{*}{\textbf{Model Type}} & \multirow{3}{*}{\textbf{Model Name}}  & \multicolumn{3}{|c|}{\textbf{Intent}} & \multicolumn{3}{|c|}{\textbf{Slot}} \\
\cline{3-8}
&  & \multicolumn{1}{|c|}{\textbf{5-shot}} & \multicolumn{1}{|c|}{\textbf{10-shot}} & \multicolumn{1}{|c|}{\textbf{20-shot}}
& \multicolumn{1}{|c|}{\textbf{5-shot}} & \multicolumn{1}{|c|}{\textbf{10-shot}} & \multicolumn{1}{|c|}{\textbf{20-shot}}  \\
\cline{3-8}
&  & \multicolumn{3}{|c|}{\textbf{Macro-F1 / Accuracy}} & \multicolumn{3}{|c|}{\textbf{Macro-F1 / Accuracy}} \\
\hline

\rowcolor{Gray}Base Model & \BLSTM & 45.28 / 60.02 & 60.62 / 64.61 & 63.60 / 80.52
& 60.78 / 83.91 & 74.28  / 90.19 & 80.57 / 93.08  \\
\hline Input Level & \texttt{+feat} & 49.40 / 63.72 & 64.34 / 73.46 & 65.16 / 83.20
& \textbf{66.84} / \textbf{88.96} & 79.67 / \textbf{93.64} & 84.95 / 95.00  \\
\hline

\rowcolor{Gray}  & \texttt{+logit} & 46.01 / 58.68 & 63.51 / 77.83 & 69.22 / \textbf{89.25}
& 63.68 / 86.18 & 76.12 / 91.64  & 83.71 / 94.43 \\
\cline{2-8}

\rowcolor{Gray} \multirow{-2}{*}{Output Level}& \texttt{+hu16} & 47.22 / 56.22 & 61.83 / 68.42 & 67.40 / 84.10
& 63.37 / 85.37 & 75.67 / 91.06 & 80.85 / 93.47  \\
\hline \multirow{2}{*}{\vspace{-2.2em}\tabincell{c}{Network Module \\ Level}} & \texttt{+two} & 40.44 / 57.22 & 60.72 / 75.14 & 62.88 /
83.65
& 60.38 / 83.63 & 73.22 / 90.08 & 79.58 / 92.57  \\
\cline{2-8} & \texttt{+two+posi} & 50.90 / 74.47 & 68.69 / 84.66 & 72.43 / 85.78
& 59.59 / 83.47 & 73.62 / 89.28 & 78.94 / 92.21 \\
\cline{2-8} & \texttt{+two+neg} & 49.01 / 68.31 & 64.67 / 79.17 & 72.32 / 86.34
& 59.51 / 83.23 & 72.92 / 89.11 & 78.83 / 92.07 \\
\cline{2-8} & \texttt{+two+both} & \textbf{54.86} / \textbf{75.36} & \textbf{71.23} / \textbf{85.44} & \textbf{75.58} / 88.80
& 59.47 / 83.35 & 73.55 / 89.54 & 79.02 / 92.22 \\
\hline
\rowcolor{Gray} & \texttt{+mem} & - & - & - & 61.25 / 83.45 & 77.83 / 90.57 & 82.98 / 93.49 \\
\cline{2-8}
\rowcolor{Gray} \multirow{-2}{*}{Few-Shot Model}  & \texttt{+mem+feat} & - & - & - & 65.08 / 88.07 & \textbf{80.64} / 93.47 & \textbf{85.45} / \textbf{95.39} \\
\hline
\hline
RE Output & \REO & \multicolumn{3}{|c|}{70.31 / 68.98} & \multicolumn{3}{|c|}{42.33 / 70.79} \\
\hline
\end{tabular}
} \caption{Results on Full Few-Shot Learning Settings. For slot filling, we do not distinguish full and partial few-shot learning settings
(see Sec.~\ref{sec_datasest}).}
\label{tab_full_few}
\vspace{-1em}
\end{table*}

\cparagraph{Intent Detection} As shown in Table \ref{tab_full_few}, except for 5-shot, all approaches improve the baseline \texttt{BLSTM}.
Our network-module-level methods give the best performance because our attention module directly receives signals from the clue words in
\REs that contain more meaningful information than the \REtag itself used by other methods. We also observe that since negative \REs are
derived from positive \REs with some noises, \texttt{posi} performs better than \texttt{neg} when the amount of available data is limited.
However, \texttt{neg} is slightly better in 20-shot, possibly because negative \REs significantly outnumbers the positive ones. Besides,
\tatt alone works better than the \texttt{BLSTM} when there are sufficient data, confirming the advantage of our two-side attention
architecture.

As for other proposed methods, the output level method (\texttt{logit}) works generally better than the input level method (\texttt{feat}),
except for the 5-shot case. We believe this is due to the fewer number of \RE related parameters and the shorter distance that the gradient
needs to travel from the loss to these parameters -- both make \texttt{logit} easier to train. However, since \texttt{logit} directly
modifies the output, the final prediction is more sensitive to the insufficiently trained weights in \texttt{logit}, leading to the
inferior results in the 5-shot setting.

To compare with existing methods of combining \NN and rules, we also implement the teacher-student network~\cite{hu2016harnessing}. This
method lets the \NN learn from the posterior label distribution produced by \FOL rules in a teacher-student framework, but requires
considerable amounts of data. Therefore, although both \texttt{hu16} and \texttt{logit} operate at the output level, \texttt{logit} still
performs better than \texttt{hu16} in these few-shot settings, since \texttt{logit} is easier to train.


It can also be seen that starting from 10-shot, \texttt{two+both} significantly outperforms pure \REO. This suggests that by using our
attention loss to connect the distributional representation of the \NN and the clue words of \REs, we can generalize \RE patterns within a
\NN architecture by using a small amount of annotated data.

\cparagraph{Slot Filling}
Different from intent detection, as shown in Table \ref{tab_full_few}, our attention loss does not work for slot filling.
The reason is that the slot label of a \textbf{\emph{target word}} (the word for which we are trying to predict a slot label) is decided
mainly by the semantic meaning of the word itself, together with 0-3 phrases in the context to provide supplementary information. However,
our attention mechanism can only help in recognizing clue words in the context, which is less important than the word itself and have
already been captured by the \BLSTM, to some extent. Therefore, the attention loss and the attention related parameters are more of a
burden than a benefit. As is shown in Fig.~\ref{atis_sample}, the model recognizes \textsl{\underline{Boston}} as \emph{fromloc.city}
mainly because \textsl{\underline{Boston}} itself is a city, and its  context word \textsl{\underline{from}} may have already been captured
by the \BLSTM and our attention mechanism does not help much. By examining the attention values of \texttt{+two} trained on the full
dataset, we find that instead of marking informative context words, the attention tends to concentrate on the target word itself. This
observation further reinforces our hypothesis on the attention loss.

On the other hand, since the \REtags provide extra information, such as type, about words in the sentence, \texttt{logit} and \texttt{feat}
generally work better. However, different from intent detection, \texttt{feat} only outperforms \texttt{logit} by a margin. This is because
\texttt{feat} can use the \REtags of all words to generate better context representations through the \NN, while \texttt{logit} can only
utilize the \REtag of the target word before the final output layer. As a result, \texttt{feat} actually gathers more information from \REs
and can make better use of them than \texttt{logit}. Again, \texttt{hu16} is still outperformed by \texttt{logit}, possibly due to the
insufficient data support in this few-shot scenario. We also see that even the \texttt{BLSTM} outperforms \REO in 5-shot, indicating while
it is hard to write high-quality \RE patterns, using \REs to boost \NNs is still feasible.


\cparagraph{Summary} The amount of extra information that a \NN can utilize from the combined \REs significantly affects the resulting
performance. Thus, the attention loss methods work best for intent detection and \texttt{feat} works best for slot filling. We also see
that the improvements from \REs decreases as having more training data. This is not surprising because the implicit knowledge embedded in
the \REs are likely to have already been captured by a sufficient large annotated dataset and in this scenario using the \REs will bring in
fewer benefits.

\subsection{Partial Few-Shot Learning}
To better understand the relationship between our approach and existing few-shot learning methods, we also implement the memory network
method \cite{kaiser2017learning} which achieves good results in various few-shot datasets. We adapt their open-source code, and add their
memory module (\texttt{mem}) to our \BLSTM model.

\begin{table}
\setlength{\tabcolsep}{0.23em}
\centering
\small{
\begin{tabular}{|l|c|c|c|}

\hline
\multirow{2}{*}{\textbf{Model}}  & \multicolumn{1}{|c|}{\textbf{5-shot}} & \multicolumn{1}{|c|}{\textbf{10-shot}} & \multicolumn{1}{|c|}{\textbf{20-shot}}  \\
\cline{2-4}
 & \multicolumn{3}{|c|}{\textbf{Macro-F1 / Accuracy}}   \\
\hline
\rowcolor{Gray} \texttt{BLSTM} & 64.73 / 91.71 & 78.55 / 96.53 & 82.05 / 97.20 \\
\hline
\texttt{+hu16} & 65.22 / 91.94 & 84.49 / 96.75 & 84.80 / 97.42 \\
\hline
\rowcolor{Gray} \texttt{+two} & 65.59 / 91.04 & 77.92 / 95.52 & 81.01 / 96.86 \\
\hline
\texttt{+two+both} & 66.62 / 92.05 & 85.75 / 96.98 & \textbf{87.97} / \textbf{97.76} \\
\hline
\rowcolor{Gray} \texttt{+mem} & 67.54 / 91.83 & 82.16 / 96.75 & 84.69 / 97.42 \\
\hline
\texttt{+mem+posi} & \textbf{70.46} / \textbf{93.06} & \textbf{86.03} / \textbf{97.09} & 86.69 / 97.65 \\
\hline

\end{tabular}
} \caption{Intent Detection Results on Partial Few-Shot Learning Setting.} \label{tab_intent_few_fill} \vspace{-1em}
\end{table}

Since the memory module requires to be trained on either many few-shot classes or several classes with extra data,
we expand our full few-shot dataset for intent detection, so that the top 3 intent labels have 300 sentences (partial few-shot).

As shown in Table~\ref{tab_intent_few_fill}, \texttt{mem} works better than \BLSTM, and our attention loss can be further combined with the memory module (\texttt{mem+posi}), with even better performance. 
\texttt{hu16} also works here, but  worse than \texttt{two+both}.
Note that, the memory module requires the input sentence to have only one embedding, thus we only use one set of positive attention for combination.

As for slot filling, since we already have extra data for frequent tags in the original few-shot data (see Sec.~\ref{sec_datasest}), we use
them directly to run the memory module. As shown in the bottom of Table \ref{tab_full_few}, \texttt{mem} also improves the base \BLSTM, and
gains further boost when it is combined with \texttt{feat}\footnote{For compactness, we only combine the best method in each
task with \texttt{mem}, but others can also be combined.}.


\subsection{Full Dataset}

\begin{table}
\setlength{\tabcolsep}{0.23em}
\centering
\small{
\begin{tabular}{|l|c|c|}

\hline
\multirow{2}{*}{\textbf{Model}} & \textbf{Intent} & \textbf{Slot} \\
\cline{2-3}
  & \textbf{\scriptsize Macro-F1/Accuracy} &  \textbf{\scriptsize Macro-F1/Micro-F1} \\
\hline
\rowcolor{Gray} \texttt{BLSTM} & 92.50 / 98.77  & 85.01 / 95.47\\
\hline
\texttt{+feat} & 91.86 / 97.65 & 86.7 / 95.55\\
\hline
\rowcolor{Gray} \texttt{+logit} & 92.48 / 98.77 & 86.94 / 95.42  \\
\hline
\texttt{+hu16} & 93.09 / 98.77 & 85.74 / 95.33  \\
\hline
\rowcolor{Gray} \texttt{+two} & 93.64 / 98.88  & 84.45 / 95.05\\
\hline
\texttt{+two+both }& \textbf{96.20} / \textbf{98.99} & 85.44 / 95.27 \\
\hline
\rowcolor{Gray} \texttt{+mem} & 93.42 / 98.77 & 85.72 / 95.37\\
\hline
\texttt{+mem+posi/feat} & 94.36 / \textbf{98.99} & \textbf{87.82} / \textbf{95.90} \\
\hline
\hline
\rowcolor{Gray} L\&L16 & - / 98.43 & - / 95.98\\
\hline

\end{tabular}
}
\caption{Results on Full Dataset. The left side of `$/$' applies for intent, and the right side for slot.}
\label{tab_full}
\vspace{-1em}
\end{table}

To answer \textbf{Q2}, we also evaluate our methods on the full dataset. As seen in Table \ref{tab_full}, for intent detection, while
\texttt{two+both} still works, \texttt{feat} and \texttt{logit} no longer give improvements.
This shows that since both \REtag and annotated data provide intent labels for the input sentence, the value of the extra noisy tag from \RE become limited as we have more annotated data.
However, as there is no guidance on attention in the annotations, the clue words from \REs are still useful. Further, since \texttt{feat}
concatenates \REtags at the input level, the powerful \NN makes it more likely to overfit than \texttt{logit}, therefore \texttt{feat}
performs even worse when compared to the \BLSTM.

As for slot filling, introducing \texttt{feat} and \texttt{logit} can still bring further improvements. This shows that the word type information contained in the
\REtags is still hard to be fully learned even when we have more annotated data.
Moreover, different from few-shot settings, \texttt{two+both} has a better macro-F1 score than the \texttt{BLSTM} for this task, suggesting
that better attention is still useful when the base model is properly trained.

Again, \texttt{hu16} outperforms the \texttt{BLSTM} in both tasks, showing that although the \REtags are noisy, their teacher-student
network can still distill useful information. However, \texttt{hu16} is a general framework to combine \FOL rules, which is more indirect
in transferring knowledge from rules to \NN than our methods. Therefore, it is still  inferior to attention loss in intent detection and
\texttt{feat} in slot filling, which are designed to combine \REs.

Further, \texttt{mem} generally works in this setting, and can receive further improvement
by combining our fusion methods.
We can also see that \texttt{two+both} works clearly better than the state-of-art method (\LL) in intent detection, which jointly models the two tasks. And \texttt{mem+feat} is comparative to \LL in slot filling.

\subsection{Impact of the RE Complexity}
\label{sec_complexity}
\begin{table}
\setlength{\tabcolsep}{0.23em}
\centering
\small{
\begin{tabular}{|l|c|c|c|c|}

\hline
\multirow{3}{*}{\textbf{Model}}  & \multicolumn{2}{|c|}{\textbf{Intent}} & \multicolumn{2}{|c|}{\textbf{Slot}}  \\
\cline{2-5}
  & \multicolumn{2}{|c|}{\textbf{\scriptsize Macro-F1 / Accuracy}} & \multicolumn{2}{|c|}{\textbf{\scriptsize Macro-F1 / Micro-F1}}  \\
\cline{2-5}
  & \textbf{\scriptsize Complex} & \textbf{\scriptsize Simple} & \textbf{\scriptsize Complex} & \textbf{\scriptsize Simple} \\
\hline
\rowcolor{Gray} \texttt{BLSTM} & \multicolumn{2}{|c|}{63.60 / 80.52} & \multicolumn{2}{|c|}{80.57 / 93.08}  \\
\hline
\texttt{+feat }& 65.16/\textbf{83.20} & \textbf{66.51}/80.40 & \textbf{84.95/95.00} & 83.88/94.71 \\
\hline
\rowcolor{Gray} \texttt{+logit} & \textbf{69.22/89.25} & 65.09/83.09 & \textbf{83.71/94.43} & 83.22/93.94  \\
\hline
\texttt{+both} & \textbf{75.58/88.80} & 74.51/87.46 & - & - \\
\hline
\end{tabular}
} \caption{Results on 20-Shot Data with Simple \REs. \texttt{+both} refers to \ptatt\texttt{+both} for short.} \label{tab_simple}
\vspace{-1em}
\end{table}

We now discuss how the \RE complexity affects the performance of the combination.
We choose to control the \RE complexity by modifying the number of groups.
%
Specifically, we reduce the number of groups for existing \REs to decrease \RE complexity.
To mimic the process of writing simple \REs from scratch, we try our best to keep the key \RE groups.
For intent detection, all the \REs are reduced to at most 2 groups. 
As for slot filling, we also reduce the \REs to at most 2 groups, and for some simples case, we further reduce them into  word-list patterns, e.g., \texttt{(\_\_CITY)}.

As shown in Table \ref{tab_simple}, the simple \REs already deliver clear improvements to the base \NN models, which shows the
effectiveness of our methods, and indicates that simple \REs are quite cost-efficient since these simple \REs only contain 1-2 \RE groups
and thus very easy to produce. We can also see that using complex \REs generally leads to better results compared to using simple \REs.
This indicates that when considering using \REs to improve a \NN model, we can start with simple \REs, and gradually increase the \RE
complexity to improve the performance over time\footnote{We do not include results of \texttt{both} for slot filling since its \REs are
different from \texttt{feat} and \texttt{logit}, and we have already shown that the attention loss method does not work for slot filling.}.



\section{Related Work}
Our work builds upon the following techniques, while qualitatively differing from each

 \cparagraph{NN with Rules}
On the initialization side, Li et al.~\shortcite{li2017initializing} uses important n-grams to initialize the convolution filters.
On the input side, Wang et al.~\shortcite{wang2017combining} uses knowledge base rules to find relevant concepts for short texts to augment input.
On the output side,
Hu et al.~\shortcite{hu2016harnessing,hu2016deep} and Guo et al.~\shortcite{guo2017knowledge} use \FOL rules to rectify the output probability of \NN, and then let \NN learn from the rectified distribution in a teacher-student framework.
Xiao et al.~\shortcite{xiao2017symbolic}, on the other hand, modifies the decoding score of \NN by multiplying a weight derived from rules.
On the loss function side, people modify the loss function to model the relationship between premise and conclusion~\cite{demeester2016lifted}, and fit both human-annotated and rule-annotated labels~\cite{alashkar2017examples}.
Since fusing in initialization or in loss function often require special properties of the task, these approaches are not applicable to our
problem. Our work thus offers new ways to exploit \RE rules at different levels of a \NN.

\cparagraph{NNs and REs} As for \NNs and \REs, previous work has tried to use \RE to speed up the decoding phase of a
\NN~\cite{strauss2016regular} and generating \REs from natural language specifications of the \RE~\cite{locascio2016neural}. By contrast,
our work aims to use \REs to improve the prediction ability of a \NN.

\cparagraph{Few-Shot Learning}
Prior work either considers few-shot learning in a metric learning framework~\cite{koch2015siamese,vinyals2016matching}, or stores
instances in a memory~\cite{santoro2016meta, kaiser2017learning} to match similar instances in the future.
Wang et al.~\shortcite{wang2017multi} further uses the semantic meaning of the class name itself to provide extra information for few-shot
learning. Unlike these previous studies, we seek to use the human-generated \REs to provide additional information.

\cparagraph{Natural Language Understanding} Recurrent neural networks are proven to be effective in both intent
detection~\cite{ravuri2015comparative} and slot filling~\cite{mesnil2015using}.
Researchers also find ways to jointly model the two tasks~\cite{liu2016attention, zhang2016joint}. However, no work so far has combined
\REs and \NNs to improve intent detection and slot filling. 

\section{Conclusions}
\vspace{-1mm} In this paper, we investigate different ways to combine \NNs and \REs for solving typical \SLU tasks. Our experiments
demonstrate that the combination clearly improves the \NN performance in both the few-shot learning and the  full dataset settings. We show
that by exploiting the implicit knowledge encoded within \REs, one can significantly improve the learning performance. Specifically, we
observe that using \REs to guide the attention module works best for intent detection, and using \REtags as features is an effective
approach for slot filling. We provide interesting insights on how \REs of various forms can be employed to improve \NNs, showing that while
simple \REs are very cost-effective, complex \REs generally yield better results.

\section*{Acknowledgement}
\vspace{-1mm} This work is supported by the National High Technology R\&D Program of China (Grant No. 2015AA015403), the National
Natural Science Foundation of China (Grant Nos. 61672057 and 61672058); the UK Engineering and Physical Sciences Research Council (EPSRC)
under grants EP/M01567X/1 (SANDeRs) and EP/M015793/1 (DIVIDEND); and the Royal Society International Collaboration Grant (IE161012). For
any correspondence, please contact Yansong Feng.


\bibliography{acl2018}
\bibliographystyle{acl_natbib}
\balance
\end{document}